\documentclass{article}

\usepackage{microtype}
\usepackage{graphicx}
\usepackage{subfigure}
\usepackage{booktabs} 

\usepackage{hyperref}

\usepackage[accepted]{synsml2023}

\usepackage{amsmath}
\usepackage{amssymb}
\usepackage{mathtools}
\usepackage{amsthm}
\usepackage{dsfont}
\newcommand\E{\mathbb E}
\usepackage{multirow}

\usepackage[capitalize,noabbrev]{cleveref}

\theoremstyle{plain}

\theoremstyle{definition}

\theoremstyle{remark}

\usepackage[textsize=tiny]{todonotes}

\synsmltitlerunning{ChemGymRL: An Interactive Framework for Reinforcement Learning for Digital Chemistry}

\begin{document}

\twocolumn[
\synsmltitle{ChemGymRL: An Interactive Framework for Reinforcement Learning for Digital Chemistry}



\synsmlsetsymbol{equal}{*}

\begin{synsmlauthorlist}
    \synsmlauthor{Chris Beeler}{uom,uw,nrc}
    \synsmlauthor{Sriram Ganapathi Subramanian}{uw,vec}
    \synsmlauthor{Kyle Sprague}{uw}
    \synsmlauthor{Nouha Chatti}{uw}
    \synsmlauthor{Colin Bellinger}{nrc}
    \synsmlauthor{Mitchell Shahen}{uw}
    \synsmlauthor{Nicholas Paquin}{uw}
    \synsmlauthor{Mark Baula}{uw}
    \synsmlauthor{Amanuel Dawit}{uw}
    \synsmlauthor{Zihan Yang}{uw}
    \synsmlauthor{Xinkai Li}{uw}
    \synsmlauthor{Mark Crowley}{uw}
    \synsmlauthor{Isaac Tamblyn}{vec,uop}
\end{synsmlauthorlist}

\synsmlaffiliation{uom}{Department of Mathematics and Statistics, University of Ottawa, Ottawa ON, Canada}
\synsmlaffiliation{uw}{Department of Electrical and Computer Engineering, University of Waterloo, Waterloo ON, Canada}
\synsmlaffiliation{nrc}{Digital Technologies, National Research Council of Canada, Ottawa ON, Canada}
\synsmlaffiliation{vec}{Vector Institute for Artificial Intelligence, Toronto ON, Canada}
\synsmlaffiliation{uop}{Department of Physics, University of Ottawa, Ottawa ON, Canada}

\synsmlcorrespondingauthor{Chris Beeler}{christopher.beeler@uottawa.ca}
\synsmlcorrespondingauthor{Mark Crowley}{mark.crowley@uwaterloo.ca}
\synsmlcorrespondingauthor{Isaac Tamblyn}{isaac.tamblyn@uottawa.ca}

\synsmlkeywords{Machine Learning}

\vskip 0.3in
]



\printAffiliationsAndNotice{}  

\begin{abstract}
This paper provides a simulated laboratory for making use of Reinforcement Learning (RL) for chemical discovery. Since RL is fairly data intensive, training agents `on-the-fly' by taking actions in the real world is infeasible and possibly dangerous. Moreover, chemical processing and discovery involves challenges which are not commonly found in RL benchmarks and therefore offer a rich space to work in. We introduce a set of highly customizable and open-source RL environments, \textbf{ChemGymRL}, based on the standard Open AI Gym template. ChemGymRL supports a series of interconnected virtual chemical \emph{benches} where RL agents can operate and train. The paper introduces and details each of these benches using well-known chemical reactions as illustrative examples, and trains a set of standard RL algorithms in each of these benches. Finally, discussion and comparison of the performances of several standard RL methods are provided in addition to a list of directions for future work as a vision for the further development and usage of ChemGymRL.
\end{abstract}

\section{Introduction}
In Material Design, the goal is to determine a pathway of chemical and physical manipulation that can be performed on some starting materials or substances in order to transform them into a desired target material. The aim of this research is to improve aspects of this process using reinforcement learning (RL). Here, we introduce a collection of interconnected RL environments called \textbf{ChemGymRL} using the \texttt{OpenAI-Gym} standard which are designed for experimentation and exploration with RL within the context of discovery and optimization of chemical synthesis. These environments are each a virtual variant of a chemistry "bench", an experiment or process that would otherwise be performed in real-world chemistry labs.

Recently, there has been a growth in research using automated robotics for chemical laboratories \cite{jiang2022artificial, she2022robotic, caramelli2021discovering, fakhruldeen2022archemist, rooney2022self, hickman2023self, bennett2022autonomous, porwol2020autonomous} and research into improving/studying these automated systems \cite{manzano2022autonomous, macleod2022self, macleod2022flexible, seifrid2022autonomous, flores2020materials}. Additionally, there has been a lot of progress on developing/using digital chemistry as a way to accelerate research for material/drug discovery \cite{m2023digitizing, bubliauskas2022digitizing, pizzuto2022solis, pyzer2021accelerating, li2021combining, fievez2022machine, yoshikawa2023digital, choubisa2023accelerated, roch2020chemos}, some even using RL \cite{volk2023alphaflow, zhou2017optimizing, gottipati2020learning}. Even with these advancements, the biggest roadblock for automated laboratories is the access to data to train them on \cite{nature2023research}.

The goal of ChemGymRL is to simulate enough of complexity of real-world chemistry experiments to allow meaningful exploration of algorithms for learning policies to control bench-specific agents, while keeping it simple enough that episodes can be rapidly generated during the RL algorithm development process. The environment supports the training of RL agents by associating positive and negative rewards based on the procedure and outcomes of actions taken by the agents. The aim is for ChemGymRL to help bridge the gap between autonomous laboratories and digital chemistry.

\begin{figure*}
    \centering
    \includegraphics[width=\textwidth]{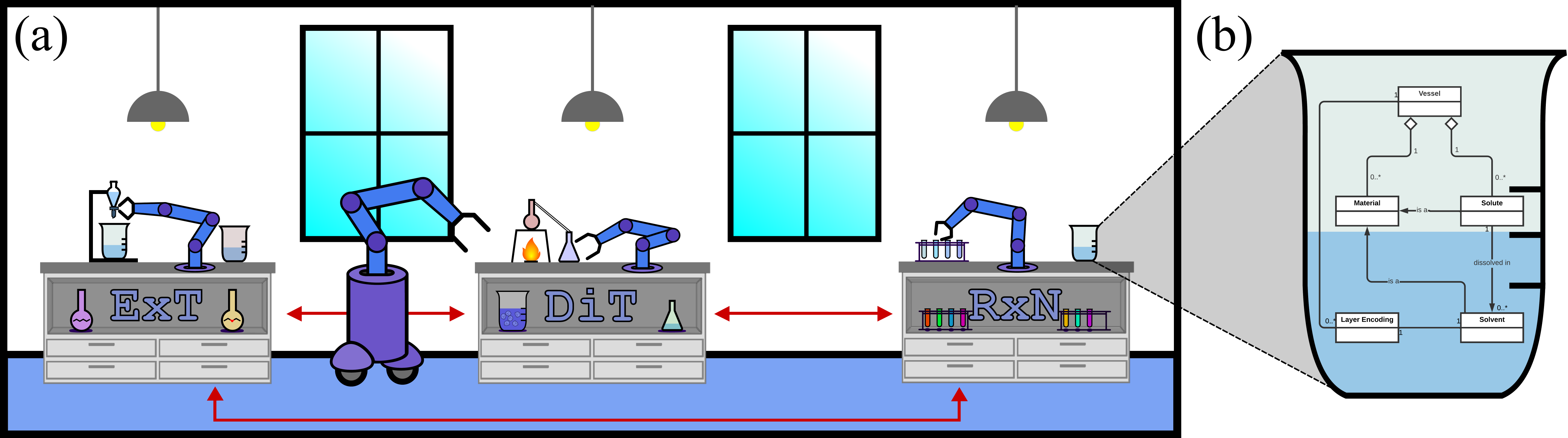}
    \caption{(a) The ChemGymRL environment. Individual agents operate at each bench, working towards their own goals. The benches pictured are extraction (\texttt{ExT}), distillation (\texttt{DiT}), and reaction (\texttt{RxN}). The user determines which materials the bench agents have access to and what vessels they start with. Vessels can move between benches; the output of one bench becomes an input of another, just as in a real chemistry lab. Successfully making a material requires the skilled operation of each individual bench as well as using them as a collective. (b) Materials within a laboratory environment are stored and transported between benches within a vessel. Benches can act on these vessels by combining them, adding materials to them, allowing for a reaction to occur, observing them (thereby producing a measurement), etc.}
    \label{fig:overview}
\end{figure*}

The ChemGymRL Open Source Library enables the use of RL algorithms to train agents in the pursuit of achieving an optimal path for generating a suitable outcome by means of a comprehensive reward system and numerous avenues for achieving an outcome.

Our goal is to create an environment that allows scientists and researchers to simulate chemical laboratories for the development of RL agents. For detailed information about this software package, documentation and tutorials including code and videos can be found at \url{https://www.chemgymrl.com/}.

We would like to specifically highlight ChemGymRL as a unique testbed for RL research. Since ChemGymRL is open source and highly customizable, it provides a training environment to accelerate RL research in several directions (especially pertaining to different RL sub-areas like hierarchical RL \cite{sutton1999}, model-based RL \cite{moerland2023model}, distributional RL \cite{bellemare2017distributional}, curriculum learning \cite{narvekar2020curriculum}, constrained RL \cite{liu2021policy}, inverse RL \cite{ng2000algorithms} etc.), in addition to providing a useful training environment with real-world applications (more discussions in Section~\ref{sec:rltestbed}). While the majority of prior RL environments have focused on computer game domains with specific challenges for the RL community\footnote{For example Sokoban-RL \cite{shoham2021solving} pertains to model-based RL, MineRL \cite{guss2019minerl} corresponds to curriculum learning etc., however ChemGymRL is a general testbed that can support a wide-variety of RL paradigms and algorithms.}, ChemGymRL pertains to a comparatively large space of RL challenges since it is associated with a real-world problem having open world difficulties. From the perspective of an RL researcher, there is a critical need for new test beds based on real-world applications that can test the limits of modern RL algorithms, which will accelerate the development of RL. We feel that ChemGymRL will be highly beneficial to the RL community in this context.  

\section{ChemGymRL}
 
\subsection{The Laboratory}
ChemGymRL is designed in a modular fashion so that new benches can be added or modified with minimal difficulty or changes to the source code. The environment can be thought of as a virtual chemistry laboratory consisting of different stations or benches where a variety of tasks can be completed, represented in Fig. \ref{fig:overview}(a). The laboratory is comprised of 3 basic elements: vessels, shelves, and benches. \textit{Vessels} contain materials, in pure or mixed form, with each vessel tracking the hidden internal state of their contents, as shown in Fig. \ref{fig:overview}(b). Whether an agent can determine this state, through measurement or reasoning, is up to the design of each bench and the user's goals. A \textit{shelf}, can hold all any \textit{vessels} not currently in use, as well as the results (or output vessels) of previous experiments to allocate vessels to benches and bench agents.

Finally, each \textit{chemistry bench} recreates a simplified version of one task in a material design pipeline. A bench must be able to receive a set of initial experimental supplies, possibly including vessels, and return the results of the intended experiment, also including modified vessels. The details and methods of how the benches interact with the vessels between these two points are completely up to the user, including the goal of the bench. In this initial version of ChemGymRL we have defined our own initial core benches, which we describe in the following sections and which will allow us to demonstrate an example workflow. 

\begin{figure}
    \centering
    \includegraphics[width=0.45\textwidth]{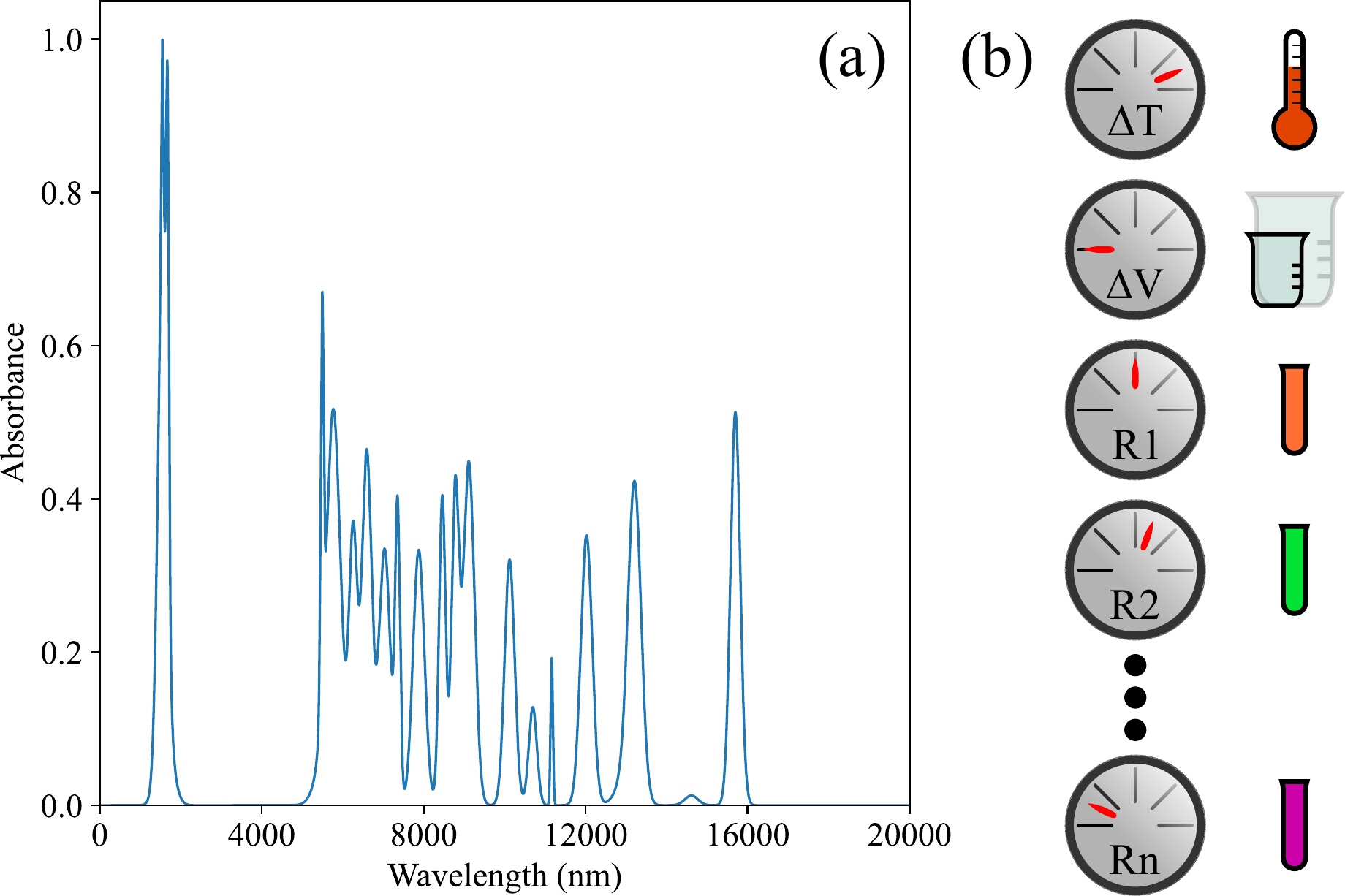}
    \caption{A visualization of the reaction bench (\texttt{RxN}) observation and action space. (a) An example of a UV-Vis spectra that would been seen in an observation and (b) The icons representing each action in \texttt{RxN}.}
    \label{fig:react}
\end{figure}

\subsection{Reaction Bench (\texttt{RxN})}
The sole purpose of the \textbf{reaction bench (\texttt{RxN})} is to allow the agent to transform available reactants into various products via a chemical reaction. The agent has the ability to control temperature and pressure of the vessel as well as the amounts of reactants added. The mechanics of this bench are quite simple in comparison to real-life which enables low computational cost for RL training. Reactions are modelled by solving a system of differential equations which define the rates of changes in concentration (See Appendix). 

\begin{figure*}
    \centering
    \includegraphics[width=\textwidth]{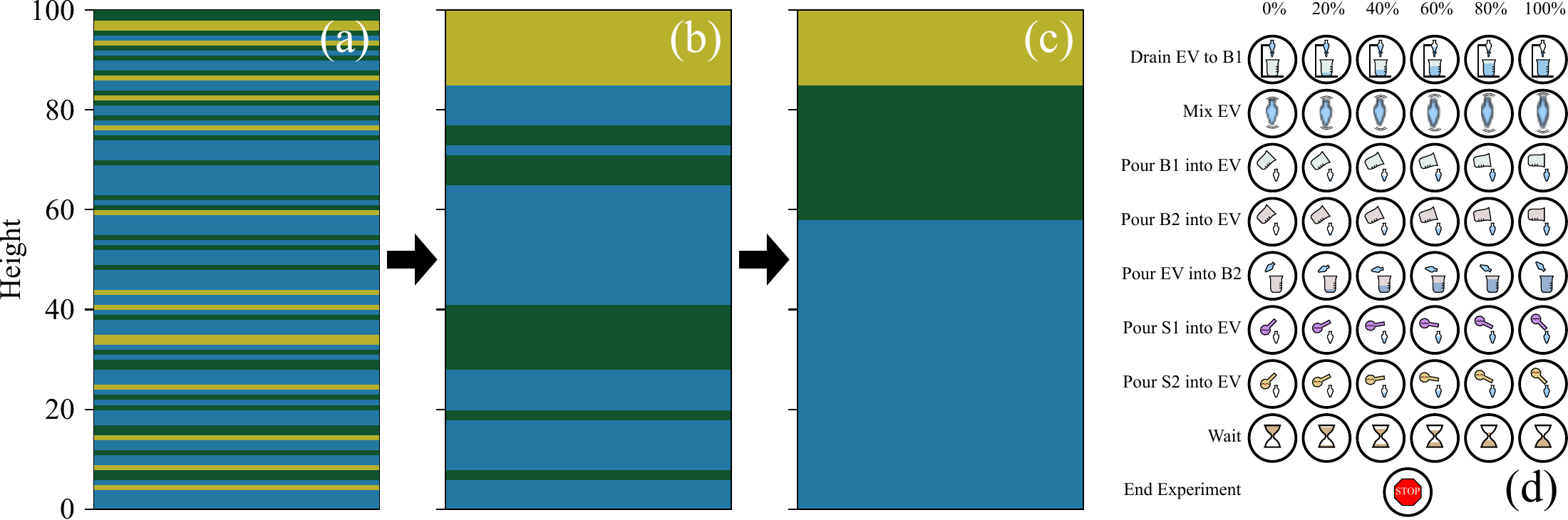}
    \caption{Typical observations seen in extraction bench (\texttt{ExT}) for a vessel containing air, hexane, and water. (a) The vessel in a fully mixed state. Each material is uniformly distributed throughout the vessel with little to no distinct layers formed. (b) The vessel in a partially mixed state. The air has formed a single layer at the top of the vessel and some distinct water and hexane layers have formed, however they are still mixed with each other. (c) The vessel in a fully settled state. Three distinct layers have formed in order of increasing density: water, hexane, and then air. (d) The icons representing each action and their multiplier values available in \texttt{ExT}. The extraction vessel (EV) is the primary vessel used, B1/B2 are the auxiliary vessels used in the experiment, and S1/S2 are the solvents available.}
    \label{fig:extractbench}
\end{figure*}

The goal of the agent operating on this bench is to modify the reaction parameters, in order to increase and/or decrease the yield of certain desired/undesired materials. The key to the agent’s success in this bench is learning how best to allow certain reactions to occur such that the yield of the desired material is maximized and the yield of the undesired material is minimized. 

\textbf{\textit{Observation Space:}} In this bench, the agent is able to observe a UV-Vis absorption spectra of the materials present in the vessel as shown in Fig. \ref{fig:react}(a), the normalized temperature, volume, pressure, and available materials for the system.

\textbf{\textit{Action Space:}} The agent can increase or decrease the temperature and volume of the vessel, as well as add any fraction of the remaining reactants available to it. In this bench, the actions returned by an agent are a continuous valued vector of size $n+2$, where $n$ is the number of reactants. These actions are also shown in Fig. \ref{fig:react}(b).

A main feature of ChemGymRL is its modularity. If one wanted to make the results of \texttt{RxN} more accurate and generalizable, they could replace the current system of differential equations with a molecular dynamics simulation without needing to change how the agent interacts with the bench or how the bench interacts with the rest of ChemGymRL.

\subsection{Extraction Bench (\texttt{ExT})}
Chemical reactions commonly result in a mixture of desired and undesired products. Extraction is a method to separate them. The extraction bench (\texttt{ExT}) aims to isolate and extract certain dissolved materials from an input vessel containing multiple materials through the use of various insoluble solvents. This is done by means of transferring materials between a number of vessels and utilizing specifically selected solvents to separate materials from each other.

A simple extraction experiment example is extracting salt from an oil solvent using water. Suppose we have a vessel containing sodium chloride dissolved in hexane. Water is added to vessel and the vessel is shaken to mix the two solvents. When the contents of the vessel settle, the water and hexane will have separated into two different layers. Sodium chloride is an ionic compound, therefore there is a distinct separation of charges when dissolved. Due to hexane being a non-polar solvent and water being a polar solvent, a large portion of the dissolved sodium chloride is pulled from the hexane into the water. Since water has a higher density than hexane, it is found at the bottom of the vessel and can be easily drained away, bringing the dissolved sodium chloride with it.

\textit{\textbf{Observation Space:}} For a visual representation of the solvent layers in the vessel for the agent, as seen in Fig. \ref{fig:extractbench}(a)-(c), we sample each solvent corresponding to each pixel using the relative heights of these distributions as probabilities. This representation makes this bench a partially observable Markov decision process (POMDP). The true state is not observed because the observations do not show the amount of dissolved solutes present nor their distribution throughout the solvents. In this set up, very light solvents will quickly settle at the top of the vessel, while very dense solvents will quickly settle at the bottom. The more similar two solvents densities are, the longer they will take to fully separate. 

\textit{\textbf{Action Space:}} The agent here has the ability to mix the vessel or let it settle, add various solvents to the vessel, drain the contents of the vessel into an auxiliary vessel bottom first, pour the contents of the vessel into a secondary auxiliary vessel, and pour the contents of either auxiliary vessel into each other or back into the original vessel. Unlike in \texttt{RxN}, only one of these actions can be performed at once, therefore the actions returned by the agent conceptually consist of two discrete values. The first value determines \textit{which} of the processes are performed and the second value determines the magnitude of that process. If the drain action is selected by the first value, then the second value determines how much is drained form the vessel. Including the ability to end the experiment, the agent has access to 8 actions with 5 action multiplier values each. These actions are depicted in Fig. \ref{fig:extractbench}(d). Practically however, the actions returned by the agent consist of a single discrete values to reduce redundancy in the action space.

The goal of the agent in this bench is to use these processes in order to maximize the purity of a desired \textit{solute} relative to other \textit{solutes} in the vessel. This means the agent must isolate the desired solute in one vessel, while separating any other solutes into the other vessels. Note that the solute's relative purity \textit{is not} affected by the presence of solvents, only the presence of other solutes.

Similar to \texttt{RxN}, if one wanted to make the results of \texttt{ExT} more realistic, they could replace the separation equations with a fluid dynamics simulation without needing to change how the agent interacts with the bench or how the bench interacts with the rest of ChemGymRL.

\subsection{Distillation Bench (\texttt{DiT})}
Similar to the \texttt{ExT}, the distillation bench (\texttt{DiT}) aims to isolate certain materials from an inputted vessel containing multiple materials (albeit with a different process). This is done by means of transferring materials between a number of vessels and heating/cooling the vessel to separate materials from each other.

A simple distillation example is extracting a solute dissolved in a single solvent. Suppose we have a vessel containing sodium chloride dissolved in water. If we heat the vessel to $100^{\circ}$C, the water will begin to boil. With any added heat, more water will evaporate and be collected in an auxiliary vessel, leaving the dissolved sodium chloride behind to precipitate out as solid sodium chloride in the original vessel.

\textit{\textbf{Observation Space:}} For a visual representation for the agent, we use the same approach described for \texttt{ExT}. For the precipitation of any solutes, we define a precipitation reaction and use the same approach described for \texttt{RxN}.

\begin{figure}
    \centering
    \includegraphics[width=0.45\textwidth]{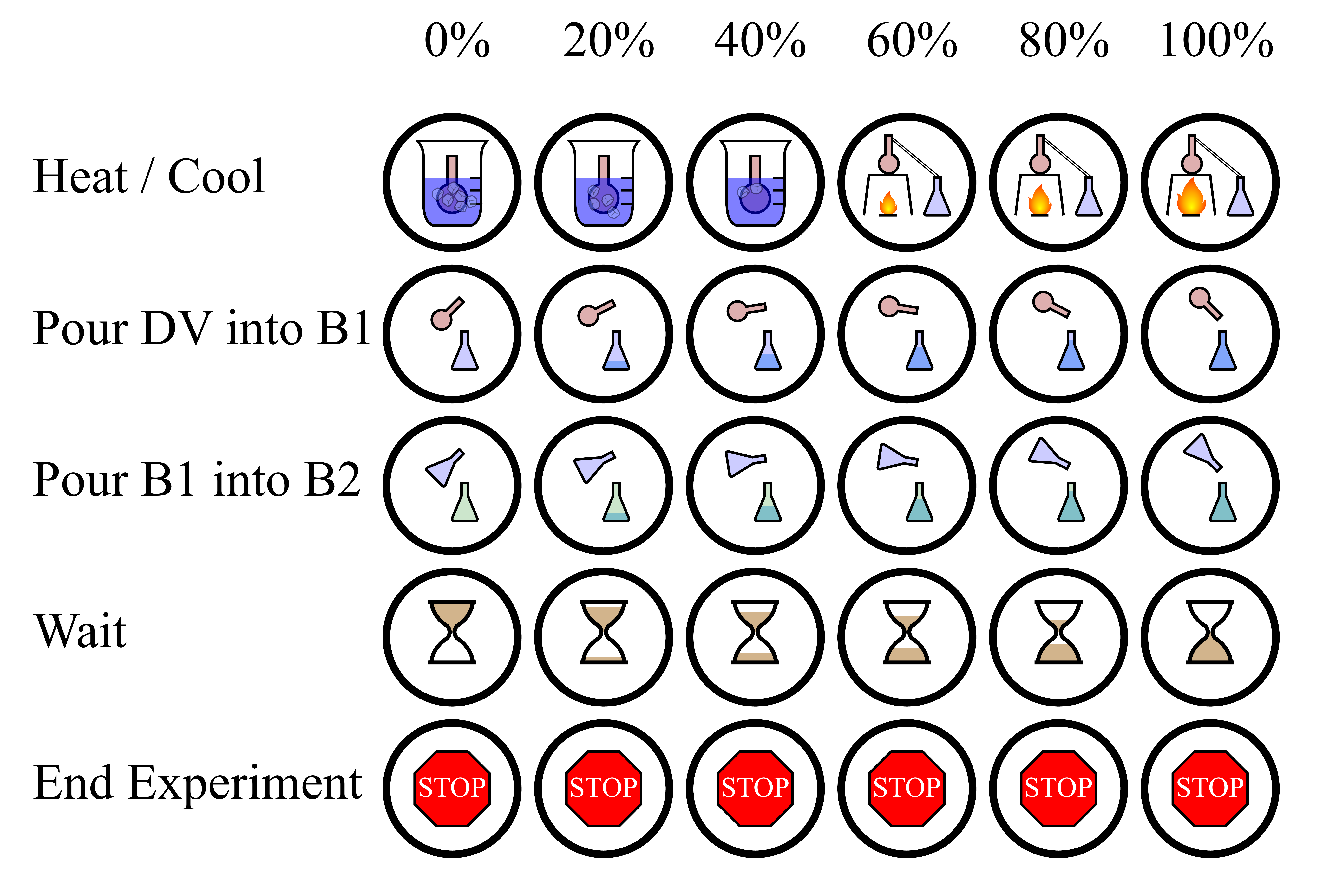}
    \caption{The icons representing each action and their multiplier values available in \texttt{DiT}. The distillation vessel (DV) is the primary vessel and B1/B2 are the auxiliary vessels in the experiment.}
    \label{fig:distill_actions}
\end{figure}

\textit{\textbf{Action Space:}} The agent has the ability to heat the vessel or let it cool down and pour the contents of any of the vessels (original and auxiliaries) into one another. When the agent heats/cools the vessel, the temperature of the vessel and its materials are altered by
\begin{equation}
    \Delta T = \frac{Q}{C},
\end{equation}
where $Q$ is the amount of heat added and $C$ is the total heat capacity  of the contents of the vessel. However, if the temperature of the vessel is at the boiling point of one of its materials, the temperature no longer increases. Instead, any heat added is used to vaporize that material according to
\begin{equation}
    \Delta n_{l} = \frac{Q}{\Delta H_{v}},
\end{equation}
where $n_{l}$ is the number of mols of the material in the liquid phase and $H_{v}$ is the enthalpy of vaporization for that material. Similar to the \texttt{ExT}, only one of these processes can be done at a time, therefore the actions returned by the agent consist of two discrete values again. Including the ability to end the experiment, the agent has access to 4 actions with 10 action multiplier values each. These actions are depicted in Fig. \ref{fig:distill_actions}. Again, the actions returned by the agent consist of a single discrete values to reduce redundancy in the action space.

The goal of the agent in this bench is to use these processes to maximize the absolute purity of a desired material in the vessel. This means the agent must isolate the desired \textit{material} in one vessel, while separating any other \textit{materials} into other vessels. Note that unlike \texttt{ExT}, the material's absolute purity is affected by the presence of all materials.

\begin{table*}
    \centering
    \begin{tabular}{|c|c|c|c|} \hline
        Reaction & R-1 & R-2 & $\text{R}_{1}$-$\text{R}_{2}$ \\ \hline
        1 & 1-chlorohexane & 1-chlorohexane & dodecane \\ \hline
        2 & 1-chlorohexane & 2-chlorohexane & 5-methylundecane \\ \hline
        3 & 1-chlorohexane & 3-chlorohexane & 4-ethyldecane \\ \hline
        4 & 2-chlorohexane & 2-chlorohexane & 5,6-dimethyldecane \\ \hline
        5 & 2-chlorohexane & 3-chlorohexane & 4-ethyl-5-methylnonane \\ \hline
        6 & 3-chlorohexane & 3-chlorohexane & 4,5-diethyloctane \\ \hline
    \end{tabular}
    \caption{All possible Wurtz reactions involving chlorohexanes. Symmetrically equivalent entries have been removed from the table as $\text{R}_{1}$-$\text{R}_{2}$ = $\text{R}_{2}$-$\text{R}_{1}$ and 6, 5, 4-chlorohexane is equivalent to 1, 2, 3-chlorohexane, respectively.}
    \label{tab:wurtz}
\end{table*}

\subsection{Characterization Bench}
In general, it is impossible to determine the exact contents of a vessel just by looking at it. Techniques exist to help characterize the contents of a vessel, however each comes with a cost. The primary cost is the monetary cost to acquire/maintain/run the instrument used. In some cases, the sample of the vessel contents being measured is destroyed during the measurement, thus incurring a different type of cost.

The characterization bench is the primary method to obtain insight as to what the vessel contains. The purpose of the characterization bench is not to manipulate the input vessel, but to subject it to analysis techniques that observe the state of the vessel, possibly including the materials inside it and their relative quantities. This does not mean that the contents of the input vessel cannot be modified by the characterization bench. This allows an agent or user to observe vessels, determine their contents, and allocate the vessel to the necessary bench for further experimentation.

The characterization bench is the only bench that is not ``operated''. A vessel is inputted to the bench along with a characterization method and the results of said method on that vessel are returned. Currently, the characterization bench consists of a UV-Vis spectrometer that returns the UV-Vis absorption spectrum of the inputted vessel. Each material in ChemGymRL has a set of UV-Vis absorption peaks defined and the UV-Vis spectrum for a vessel is the combination of the peaks for all materials present, weighted proportionally by their concentrations. In future versions of ChemGymRL we will expand the characterization bench to include other forms of partial observation.

\section{Reinforcement Learning} 

Reinforcement Learning (RL) \cite{sutton2018reinforcement} is one possible solution to a Markov Decision Process (MDP). MDPs are represented as a tuple $\langle S, A, R, T, \gamma \rangle$ where the $s \in S \subseteq \mathds{R}^n$ denotes the state space, $a \in A \subseteq \mathds{R}^m$ denotes the action space, $r \in R \subseteq \mathds{R}$ denotes the reward function and $T = P(s_{t+1}|s_t, a_t)$ denotes the transition dynamics that provides the probability of state $s_{t+1}$ at the next time step given that the agent is in state $s_t$ and performs action $a_t$. The objective for an RL agent is to learn a policy $\pi(a|s)$ that maximizes the discounted sum of expected rewards provided by the equation $J_\pi(s) = \E_{\pi}[\sum_{t=0}^\infty \gamma^t r_t | s_0 = s$], where $\gamma \in [0,1)$ is the discount factor. 

In RL there are two broad classes of algorithms that exist in the literature. The first is $Q$-learning, where a $Q$-function is learned iteratively using the Bellman optimality operator $\mathcal{B}^* Q(s,a) = r(s,a) + \gamma \E_{s' \sim T(s'|s,a)} [\max_{a'} Q(s',a') ]$. Here $s$ and $s'$ denote the current and next state respectively, $a$ and $a'$ denotes the current and next action respectively. Finally, an exact or approximate scheme of maximization is used to extract the greedy policy from the $Q$-function. As seen from the max operator being used in the Bellman equation, this family of methods only apply to an environment with a set of discrete actions.

The second class of RL algorithms is actor-critic, where the algorithm alternates between computing a value function $Q^\pi$ by a (partial) policy evaluation routine using the Bellman operator on the stochastic policy $\pi$, and then improving the current policy $\pi$ by updating it towards selecting actions that maximize the estimate maintained by the $Q$-values. This family of methods apply to both discrete and continuous action space environments. Because of this, actor-critic may be used on any bench in chemistry gym environment.

\section{Case Study}

As a simple example, we outline how a particular chemical production process uses each of the benches.

\subsection{Wurtz Reaction}
Wurtz reactions are commonly used for the formation of certain hydrocarbons. These reactions are of the form:
\begin{equation}
    2\text{R-Cl} + 2\text{Na} \xrightarrow[]{\text{diethyl ether}} \text{R-R} + 2\text{NaCl}.
\end{equation}
Here we consider the case of hexane ($\text{C}_{6}\text{H}_{14}$) for R, where one hydrogen atoms is replaced with chlorine, giving us 1-, 2-, and 3-chlorohexane as reactants with sodium. Note that we may have 2R-Cl and R-R be replaced with $\text{R}_{1}$-Cl, $\text{R}_{2}$-Cl, and $\text{R}_{1}$-$\text{R}_{2}$ in this reaction format. Tab. \ref{tab:wurtz} shows the possible outcomes of this reaction. Note that it is impossible to produce just 5-methylundecane, 4-ethyldecane, or 4-ethyl-5-methylnonane. If the desired reaction is
\begin{equation}
    \text{R}_{1}\text{-Cl} + \text{R}_{2}\text{-Cl} + 2\text{Na} \xrightarrow[]{\text{diethyl ether}} \text{R}_{1}\text{-R}_{2} + 2\text{NaCl},
\end{equation}
then we will unavoidably also have
\begin{equation}
    \begin{aligned}
        2\text{R}_{1}\text{-Cl} + 2\text{Na} &\xrightarrow[]{\text{diethyl ether}} \text{R}_{1}\text{-R}_{1} + 2\text{NaCl} \\
        2\text{R}_{2}\text{-Cl} + 2\text{Na} &\xrightarrow[]{\text{diethyl ether}} \text{R}_{2}\text{-R}_{2} + 2\text{NaCl},
    \end{aligned}
\end{equation}
occurring simultaneously.

Wurtz can an interesting and challenging reaction because the yield varies greatly between each product, making it difficult to train an agent which can optimally make each of them.

\subsection{Workflow}

Suppose that we have the previously listed chlorohexanes, sodium, diethyl ether, and water available to us with the goal to produce dodecane. Using \texttt{RxN} we can add diethyl ether, 1-chlorohexane, and sodium to a vessel. With time, this will produce a vessel containing dodecane and sodium chloride dissolved in diethyl ether. The UV-Vis spectrometer in the \texttt{RxN} can be used to measure the progression of the reaction.

The vessel can then be brought to the \texttt{ExT} to separate dodecane from sodium chloride. Dodecane is non-polar, so if we add water to the vessel and mix, most of the sodium chloride will be extracted into the water while most of the dodecane will be left in the diethyl ether. We can then drain the water out of the vessel while keeping the diethyl ether. While it's impossible to get all of the sodium chloride out with this method, we can repeat this process to increase the purity of dodecane.

The vessel can then be brought to the \texttt{DiT} to separate the dodecane from the diethyl ether. Diethyl ether has a much lower boiling point than dodecane so it will boil first. Heating the vessel enough will cause all of the diethyl ether to vaporize, leaving the dodecane in the vessel with trace amounts of sodium chloride.

Alternatively, because dodecane has a much lower boiling point than sodium chloride, we can skip the \texttt{ExT} and bring the vessel to \texttt{DiT} right after \texttt{RxN}. As before, heating the vessel enough will cause all of the diethyl ether to vaporize, condensing into an auxiliary vessel. We can then put the collect diethyl ether elsewhere such that the auxiliary vessel collected the vaporized materials is now empty. If the vessel is heated up even further now, the dodecane will be vaporized and collected into the empty auxiliary vessel, leaving the sodium chloride behind. Our auxiliary vessel now contains pure dodecane, concluding the experiment.

While this example workflow uses the benches in a specific order, more complicated experiments may use them in a completely different order or even use each bench multiple times. Given specific goals, below we will outline how RL can be used to emulate this behavior for various cases.

\section{RL Details}

In our discrete benches, Proximal Policy Optimization (PPO) \cite{schulman2017proximal}, Advantageous Actor-Critic (A2C) \cite{Mnih2016} and Deep Q-Network (DQN) \cite{Mnih2015} were used. In our continuous benches, Soft Actor-Critic (SAC) \cite{haarnoja2018soft} and Twin Delayed Deep Deterministic Policy Gradient (TD3) \cite{fujimoto2018addressing} were used instead of DQN. Note that we can choose between SAC and TD3 fairly arbitrarily since they are both off-policy algorithms that use Q-learning.

Unless otherwise specified, all RL agents were trained for 100K time steps across 10 environments in parallel (for a total of 1M time steps). Training was done by repeatedly gathering 256 time steps of experience (in each environment), then updating our policy and/or Q-function with this new experience. Since PPO and A2C are on-policy, their policies were only updated with the 2560 steps of new experiences. In contrast, a replay buffer of size 1M was maintained and sampled when training with DQN, SAC, and TD3. For the first 30K steps of DQN training, a linear exploration schedule beginning at 1.0 and ending at 0.01 was used. Exploration remained at 0.01 afterwards. All of these RL algorithms were performed using the Stable Baselines 3 \cite{raffin2021stable} implementations.

\section{Laboratory Setup}

\subsection{Reaction Bench Methodology}
For the reaction bench (\texttt{RxN}), we consider two chemical processes. In both processes, each episode begins with a vessel containing 4 mols of diethyl ether, and operates for 20 steps. In the first process, the agent has access to 1.0 mol each of 1, 2, 3-chlorohexane, and sodium, where the system dynamics are defined by the Wurtz reaction outlined above. Each episode, a target material is specified to the agent via length 7 one-hot vector where the first 6 indices represent the 6 Wurtz reaction products in Tab. \ref{tab:wurtz} and the last represents NaCl. After the 20 steps have elapsed, the agent receives a reward equal to the molar amount of the target material produced.

In the second experiment, we introduce a new set of reaction dynamics given by
\begin{equation}
    \begin{aligned}
        A + B + C &\rightarrow E \\
        A + D &\rightarrow F \\
        B + D &\rightarrow G \\
        C + D &\rightarrow H \\
        F + G + H &\rightarrow I
    \end{aligned}
\end{equation}
where the agent has access to 1.0 mol of $A$, $B$, $C$ and 3.0 mol of $D$. Each episode, a target material is specified to the agent via length 5 one-hot vector with indices representing $E$, $F$, $G$, $H$, and $I$. If the target is $E$, the agent receives a reward equal to the molar amount of $E$ produced after the 20 steps have elapsed. Otherwise, the agent receives a reward equal to the difference in molar amounts between the target material and $E$ after the 20 steps have elapsed. Here, $E$ is an undesired material. The reaction $A + B + C \rightarrow E$ occurs quickly relative to the others, adding difficulty to the reaction when $E$ is not the target.

\begin{figure*}
    \centering
    \includegraphics[width=\textwidth]{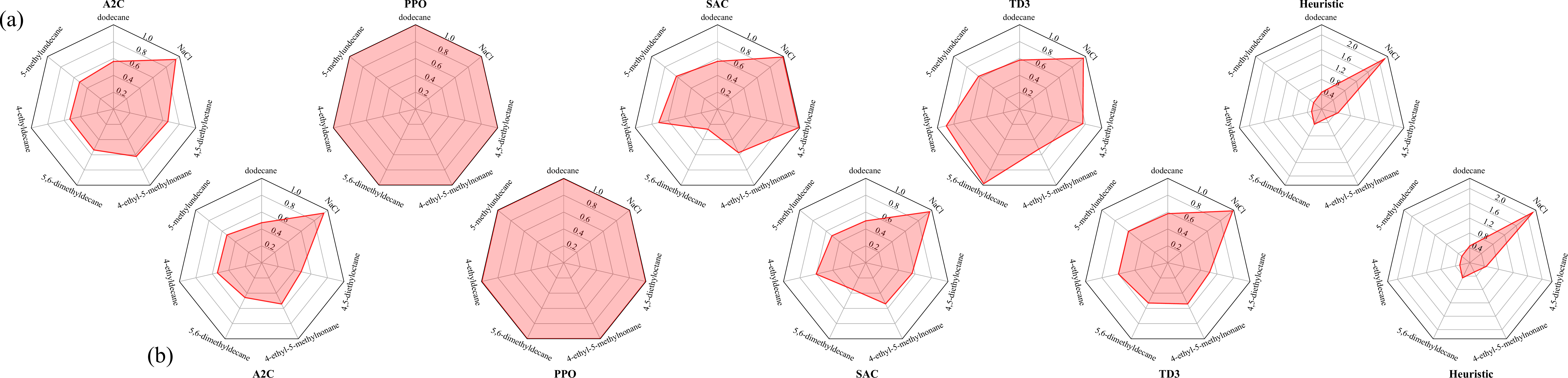}
    \caption{Radar graphs detailing the average return of each policy with respect to each target material in Wurtz \texttt{RxN}. Panel (a) uses the best policy produced from 10 runs, whereas panel (b) averages across the 10 runs (still using the best policy of each run). Returns of each RL algorithm are relative to the heuristic policy and clipped into the range $[0,\infty)$. Here, the PPO agents consistently outperform the A2C, SAC, and TD3 agents for all 7 target materials. Target materials with high returns across each algorithm (such as sodium chloride) appear to be easier tasks to learn, where target materials with less consistent return across each algorithm (such as 5,6-dimethyldecane) appear to be more difficult tasks to learn.}
    \label{fig:wurtzreact_returns}
\end{figure*}

\subsection{Extraction Bench Methodology}
For the extraction bench (\texttt{ExT}), we consider a layer-separation process where the agent operates for up to 50 steps. Similar to the Wurtz reaction, the target material is specified via length 7 one-hot vector. Each episode begins with a vessel containing 4 mols of diethyl ether, 1 mol of dissolved sodium chloride, and 1 mol of one of the 6 Wurtz reaction products in Tab. \ref{tab:wurtz}. The Wurtz reaction product contained in the vessel is the same as the target material, unless the target material is sodium chloride, in which case dodecane is added since sodium chloride is already present. After the episode has ended, the agent receives a reward equal to the change in solute purity of the target material weighted by the molar amount of that target material, where the change in solute purity is relative to the start of the experiment. If the target material is present in multiple vessels, a weighted average of the solute purity across each vessel is used.

As an example, consider when the target material is dodecane. In this experiment, the 1 mol of dissolved sodium chloride becomes 1 mol each of Na$^{+}$ and Cl$^{-}$, so the initial solute purity of dodecane is 1/3. Suppose we end the experiment with 0.7 mols of dodecane with 0.2 mols each of Na$^{+}$ and Cl$^{-}$ in one vessel, and the remaining molar amounts in a second vessel. Dodecane has a solute purity of 7/11 and 3/19 in each vessel respectively. The final solute purity of dodecane would be $0.7 \times 7/11 + 0.3 \times 3/19 \approx 0.493$. Thus the agent would receive a reward of $0.159$.

\subsection{Distillation Bench Methodology}
For the distillation bench (\texttt{DiT}), we consider a similar experimental set-up to the \texttt{ExT} one. Each episode begins with a vessel containing 4 mols of diethyl ether, 1 mol of the dissolved target material, and possibly 1 mol of another material. If the target material is sodium chloride, the additional material is dodecane, otherwise the additional material is sodium chloride. After the episode has ended, the agent receives a reward calculated similarly to the \texttt{ExT}, except using absolute purity rather than solute purity.

\section{RL Results}

\subsection{Reaction Bench}

Since reaction bench (\texttt{RxN}) has a continuous action space, we trained SAC and TD3 in addition to A2C and PPO. For the first experiment, we are looking at the Wurtz reaction dynamics. Given that we know the system dynamics in this case, we have also devised a heuristic agent for the experiment, which we expect to be optimal. This agent increases the temperature, and adds only the required reactants for the desired product immediately. This heuristic agent achieves an average return of approximately 0.62. Using the performance of this heuristic as a reference, the best and mean relative performances of the agents trained with each algorithm is shown in Fig. \ref{fig:wurtzreact_returns}. Each algorithm can consistently give rise to agents that produce sodium chloride when requested. Since this is a by-product of all reactions in our set-up, it is the easiest product to create. While the other products are not hard to produce either, they require specific reactants, and in order to maximize the yield, they require the absence of other reactants. The PPO agents are able to match the heuristic agent for all targets, while some SAC and TD3 agents are able to come close on a few targets. A2C only comes close to the heuristic on producing sodium chloride.

\begin{figure}
    \centering
    \includegraphics[width=0.45\textwidth]{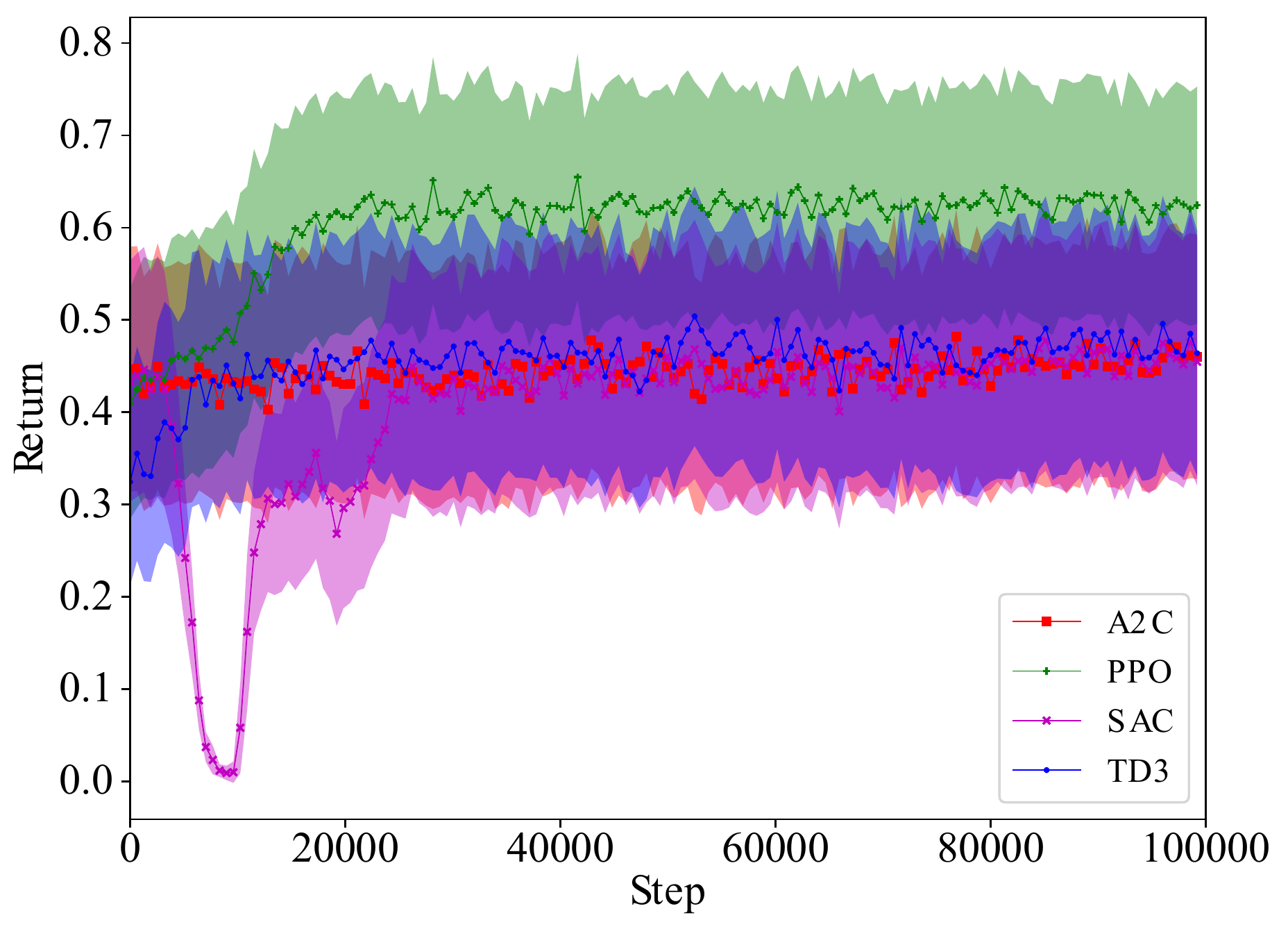}
    \caption{Wurtz \texttt{RxN}, average return with $\sigma$/5 shaded, 10 runs for each algorithm with 10 environments in parallel per run, 1M (100K sequential steps x 10 environments) total steps per run, averages are over 3200 returns. The performance of each algorithm converges before 300K total steps, with only PPO converging on an optimal policy. Despite training for an additional 700K total steps, A2C, SAC, and TD3 were not able to escape the local maxima they converged to.}
    \label{fig:wurtzreact_performance}
\end{figure}

The average return as a function of training steps for each algorithm is shown in Fig. \ref{fig:wurtzreact_performance}. On average, the agents trained with each algorithm are able to achieve a return of at least 0.4. This is expected as even an agent taking random actions can achieve an average return of approximately 0.44. The agents trained with A2C, SAC, and TD3 do not perform much better than a random agent in most cases, however the ones trained with PPO significantly outperform it. While on average, A2C, SAC, and TD3 have similar performance, we saw in Fig. \ref{fig:wurtzreact_returns} that the best performing SAC and TD3 agents outperformed the best A2C agents.

\begin{figure*}
    \centering
    \includegraphics[width=\textwidth]{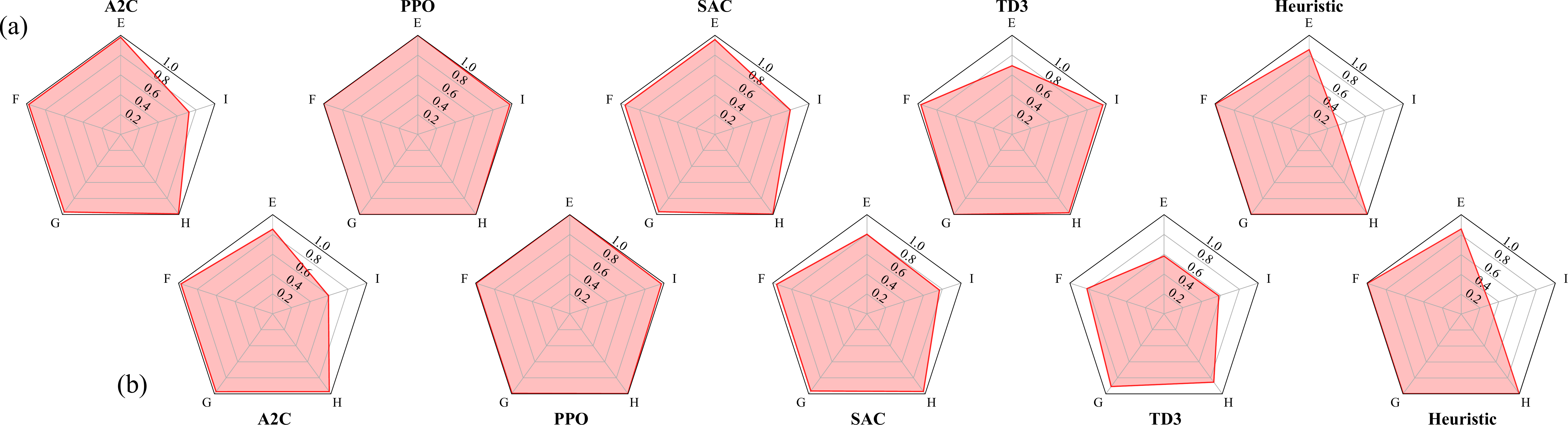}
    \caption{Radar Graph detailing the average return of each policy with respect to each target material in Fictitious \texttt{RxN}. Panel a) uses the best policy produced from 10 runs, whereas panel b) averages across the 10 runs (still using the best policy of each run). Returns of each RL algorithm are relative to the heuristic policy and clipped into the range $[0,\infty)$. Again, the PPO agents consistently outperform the A2C, SAC, and TD3 agents for all 5 target materials, however it is not as significant of a gap as in Wurtz \texttt{RxN}. Target materials with high returns across each algorithm (such as F, G, and H) appear to be easier tasks to learn, where target materials with less consistent return across each algorithm (such as E and I) appear to be more difficult tasks to learn.}
    \label{fig:fictreactv1_returns}
\end{figure*}

The second \texttt{RxN} experiment uses reaction dynamics more complicated than the Wurtz reaction. In the Wurtz reaction, the agent need only add the required reactants for the desired product all together. In this new reaction, this is still true for some desired products, however not all of them. Similarly to the previous experiment, we also devised a heuristic agent for this experiment, which achieves an average return of approximately 0.83. Using the performance of the heuristic agent as reference again, the best and mean relative performances of the agents trained with each algorithm are shown in Fig. \ref{fig:fictreactv1_returns}. Once again, PPO consistently produces agents that can match the performance of the heuristic agent. The best performing policies produced by A2C and SAC are able to nearly match the heuristic agent for all desired products excluding I. This is not unexpected as producing I requires producing intermediate products at different times during the reaction. The best performing policy produced by TD3 however, nearly matches the heuristic agent for all desired products excluding E. This is also not unexpected, given that producing E is penalized for all other desired products.

\begin{figure}
    \centering
    \includegraphics[width=0.45\textwidth]{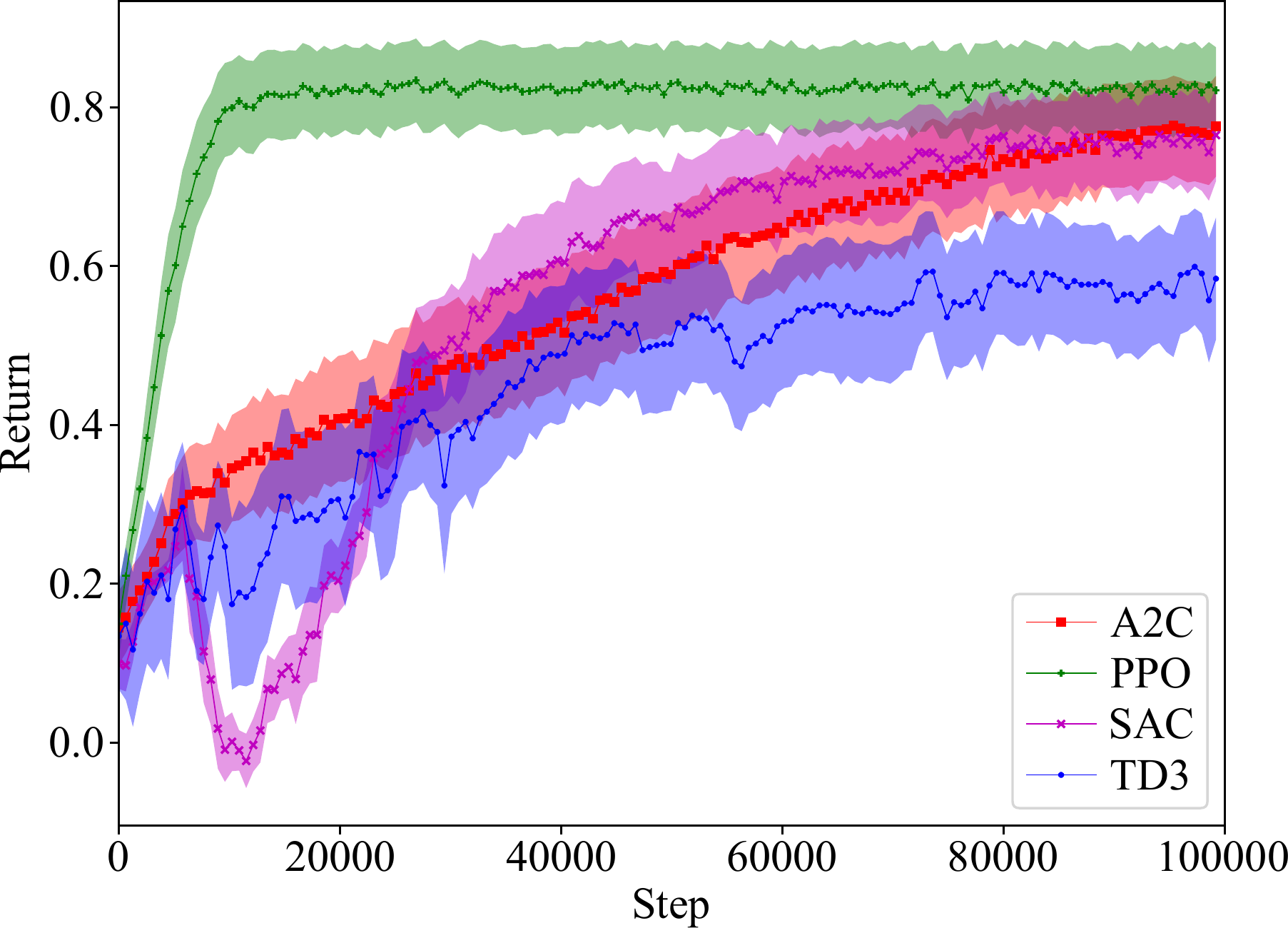}
    \caption{Fictitious \texttt{RxN}, average return with $\sigma$/5 shaded, 10 runs for each algorithm with 10 environments in parallel per run, 1M (100K sequential steps x 10 environments) total steps per run, averages are over 3200 returns. PPO quickly converges to an optimal policy, like in Wurtz \texttt{RxN}. Unlike in Wurtz \texttt{RxN}, the other algorithms take much longer to converge. While they still converge to sub-optimal performances, the gap between optimal performance is less severe.}
    \label{fig:fictreactv1_performance}
\end{figure}

Unlike PPO, the other algorithms used are less reliable at producing these best performing agents. This is likely due to PPO learning these policies much faster than the other algorithms, as seen in Fig. \ref{fig:fictreactv1_performance}. Since PPO converges to optimal behavior so quickly, there's very little room for variation in the policy. The other algorithms however are slowly converging to non-optimal behaviors, leaving much more room for variation in the policies (and returns) that they converge to.

\begin{figure*}[h!]
    \centering
    \includegraphics[width=\textwidth]{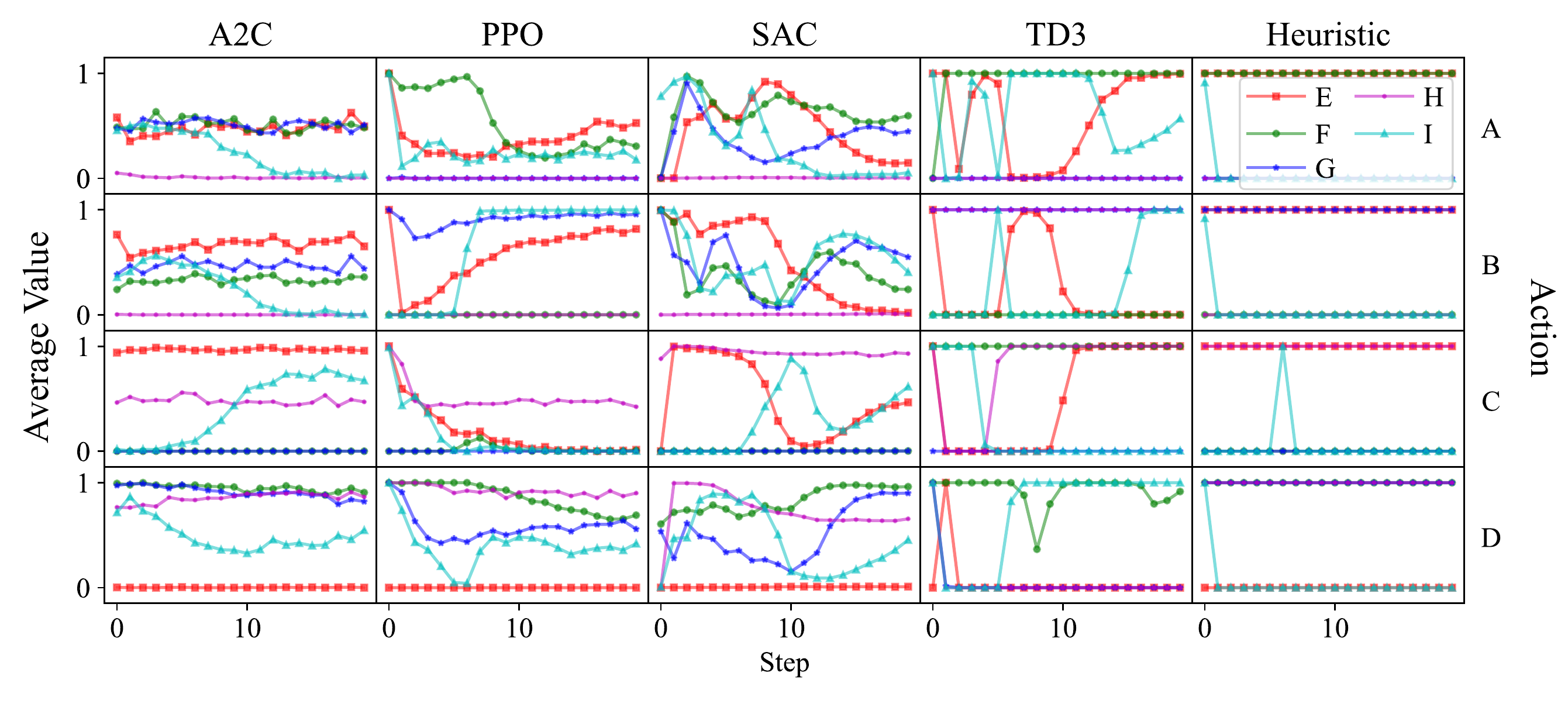}
    \caption{Fictitious \texttt{RxN}, average value of each action at every step for the best performing policies for each algorithm. The five curves in each box represents the sequence of actions for the five different target materials. Comparing the same curve across a single column outlines how a single policy acts for a single target material. Comparing different curves within a single box outlines how a single policy acts differently between different target materials. Comparing the same curve across a single row outlines how different policies act for the same target material. For actions corresponding to adding material, the curves represent how quickly those materials are added. The well performing policies are the ones that add only the required reactants (such as A2C and SAC), while the best performing policies are the ones that add them according to the right schedule (such as PPO).}
    \label{fig:fictreactv1_policies}
\end{figure*}

For the best performing agents produced by each algorithm, the average action values for each target are shown in Fig. \ref{fig:fictreactv1_policies}. Looking at the heuristic policy, a constant action can be used for each target product, excluding I. When the target is I, the desired action must change after several steps have passed, meaning the agent cannot just rely on what the specified target is. Note that if all of a material has been added by step $t$, then it does not matter what value is specified for adding that material at step $t+1$.

The best performing agent for A2C, PPO, and SAC were all able to produce E when requested and Fig. \ref{fig:fictreactv1_policies} shows that they each have learned to add A, B, C, and not D. TD3 however adds D in addition to A, B, and C, thus leading to the decreased return in that case. It can also be seen that all four algorithms learned to add two of A, B, or C in addition to D, then add the third one several steps later when I is the target product, mimicking the behavior of the heuristic policy. Note that even though the heuristic waits to add C, waiting to add A or B instead would be equally optimal. While each algorithm does this, PPO and TD3 do so better than the others. PPO is also the only one that succeeds in both of these cases, showing that an RL agent can learn the required behavior in this system.

\subsection{Extraction Bench}

With the results seen in the \texttt{RxN} tests, we now move onto the extraction bench (\texttt{ExT}) experiment. Regardless of the target material in our Wurtz extraction experiment, the optimal behavior is quite similar so we will not focus on the different cases as before. Since the \texttt{ExT} uses discrete actions, we replace SAC and TD3 with DQN. We also use what we call PPO-XL which is PPO trained with more environments in parallel. We have devised a heuristic policy for this experiment based on what an undergraduate chemist would learn. However, as the dynamics are more complex we do not necessarily expect it to be optimal.

\begin{figure}
    \centering
    \includegraphics[width=0.45\textwidth]{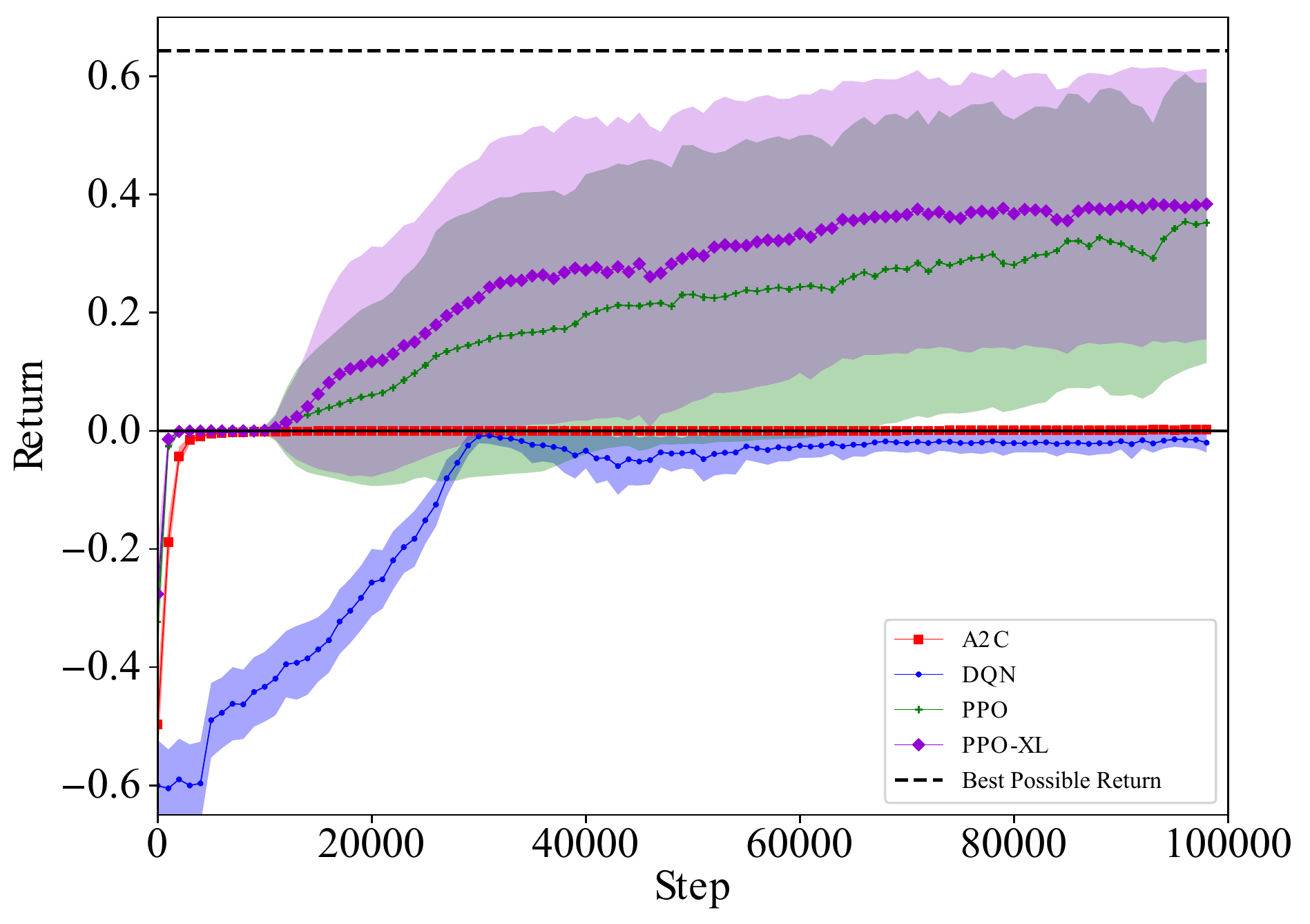}
    \caption{Wurtz \texttt{ExT}, average return with $\sigma$ shaded, 30 runs for each algorithm with 1M total steps per run (2M for PPO-XL). For each run, returns are averaged over 1000 steps (only using terminated episodes). The mean and standard deviation are then calculated across the 30 runs ($\sigma$ is calculated from 30 points). The PPO and PPO-XL agents consistently acquire positive returns, even approaching the theoretical maximum in some cases. The A2C agents learn policies which perform equivalently to ending the experiment immediately and are unable to escape those local maxima. The DQN agents acquire negative return, which is a worse performance than not running the experiment.}
    \label{fig:wurtzextract_performance}
\end{figure}

As seen in Fig. \ref{fig:wurtzextract_performance}, the agents trained with A2C do not achieve a return above zero, while the agents trained with DQN ended up achieving a negative return. Not only do both PPO and PPO-XL produce agents that achieve significantly more reward than the other algorithms, they are able to outperform the heuristic policy as well. On average, the best performing agent trained with PPO-XL manages to achieve a return of approximately 0.1 higher than the heuristic (see Fig. \ref{fig:wurtzextract_performance}), resulting in roughly a 10\% higher solute purity. While there is a large variance in the final performance of the agents trained with PPO and PPO-XL, they consistently outperform the agents trained with the other algorithms.

\begin{figure}
    \centering
    \includegraphics[width=0.45\textwidth]{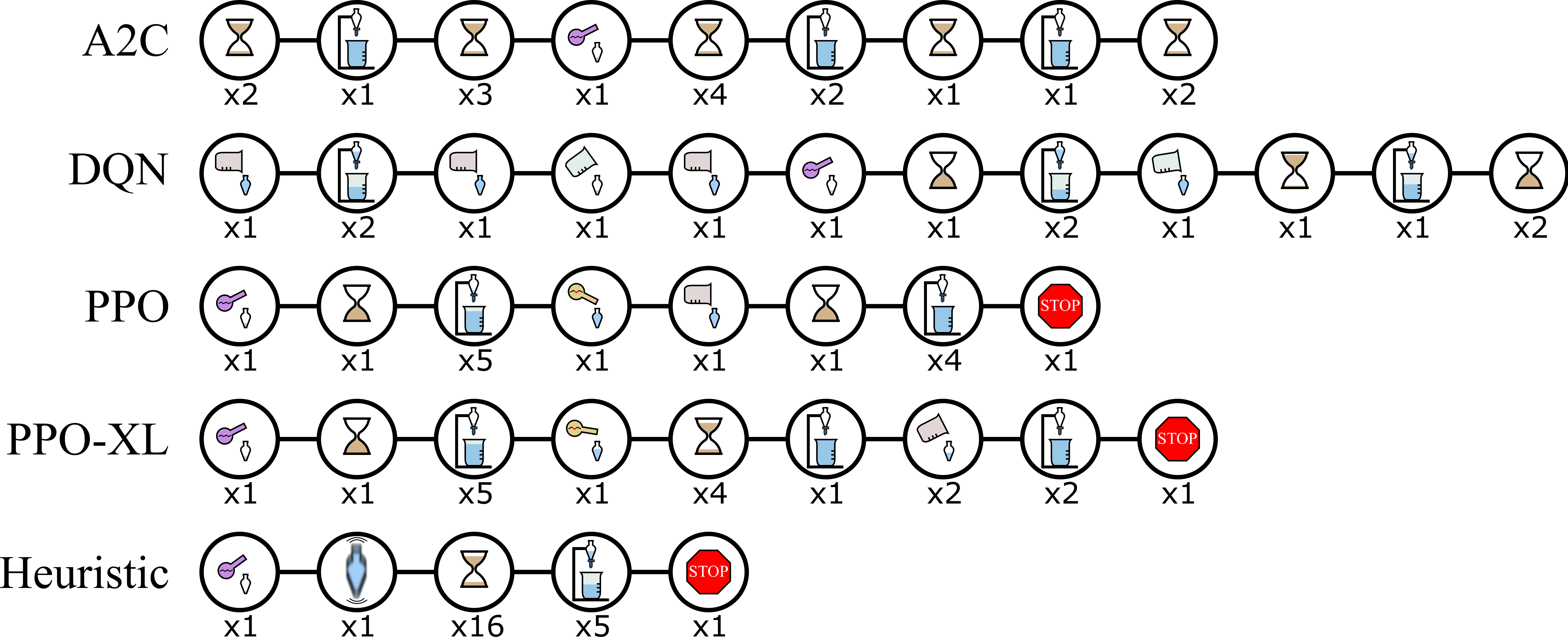}
    \caption{Wurtz \texttt{ExT}, the sequence of actions with the highest return when dodecane is the target material seen during the rollout of the best performing policy learned by each algorithm. Each picture represents an action and average value described by Fig. \ref{fig:extractbench}(d). The number beneath the image represents how many times that action was repeated. While it is more difficult to interpret these policies than with \texttt{RxN}, similarities can be seen between the PPO, PPO-XL, and heuristic policies, explaining their high performances. The A2C policy uses a similar action set, however in a different order, outlining the precision required by the agent. The DQN policy use many actions that either undo previous actions or do nothing in that specific state.}
    \label{fig:wurtzextract_policy}
\end{figure}

As shown in Fig. \ref{fig:wurtzextract_policy}, the action sequences of the policies learned from A2C, DQN, and PPO are quite different. The action sequences of the policies learned by PPO and PPO-XL are much more similar, as expected. The first half of these sequences are comparable to the heuristic, however the agents in both cases have learned a second component to the trajectory to achieve that extra return. Interestingly, both PPO and PPO-XL agents have learned to end the experiment when they achieve the desired results, whereas the A2C and DQN agents do not. PPO once again shows that an RL agent can learn the required behavior in this system.

\subsection{Distillation Bench}

\begin{figure}
    \centering
    \includegraphics[width=0.45\textwidth]{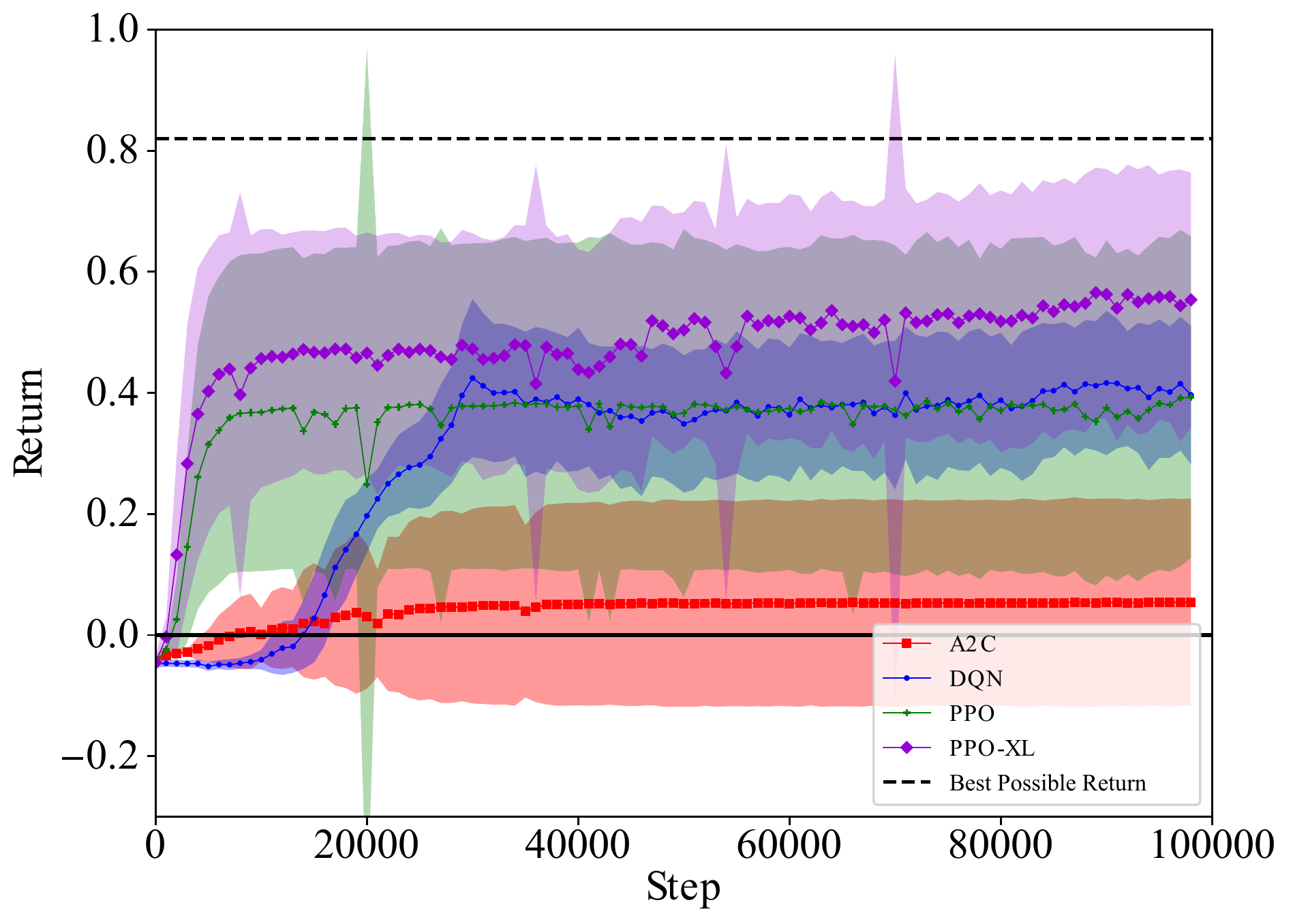}
    \caption{Wurtz \texttt{DiT}, average return with $\sigma$ shaded, 30 runs for each algorithm with 1M total steps per run (2M for PPO-XL). For each run, returns are averaged over 1000 steps (only using terminated episodes). The mean and standard deviation are then calculated across the 30 runs ($\sigma$ is calculated from 30 points). The DQN, PPO, and PPO-XL agents consistently acquire positive returns whereas the A2C agents only get positive returns on average. While DQN and PPO acquire similar returns on average, the variance with PPO is much higher, meaning the best performing PPO policy outperforms the best DQN policy. The PPO-XL policies outperform the other algorithms both on average and in the best case scenarios.}
    \label{fig:wurtzdistill_performance}
\end{figure}

Lastly, we now move onto the final experiment, distillation bench (\texttt{DiT}). Similar to \texttt{ExT}, the desired target material in the Wurtz distillation experiment does not have much effect on the optimal behavior so we will not focus on the different target cases. Instead we will focus on the different cases of when salt is and is not present with another material in the initial distillation vessel. Note that a single agent operates on both of these cases, not two agents trained independently on each case. As before, we have devised a heuristic policy and as with the \texttt{RxN} experiments, we expect it to be optimal once again. In Fig. \ref{fig:wurtzdistill_performance} we can see that on average, the algorithms converge faster than in the other experiments, however there is much more variation in the solutions.

\begin{figure}
    \centering
    \includegraphics[width=0.45\textwidth]{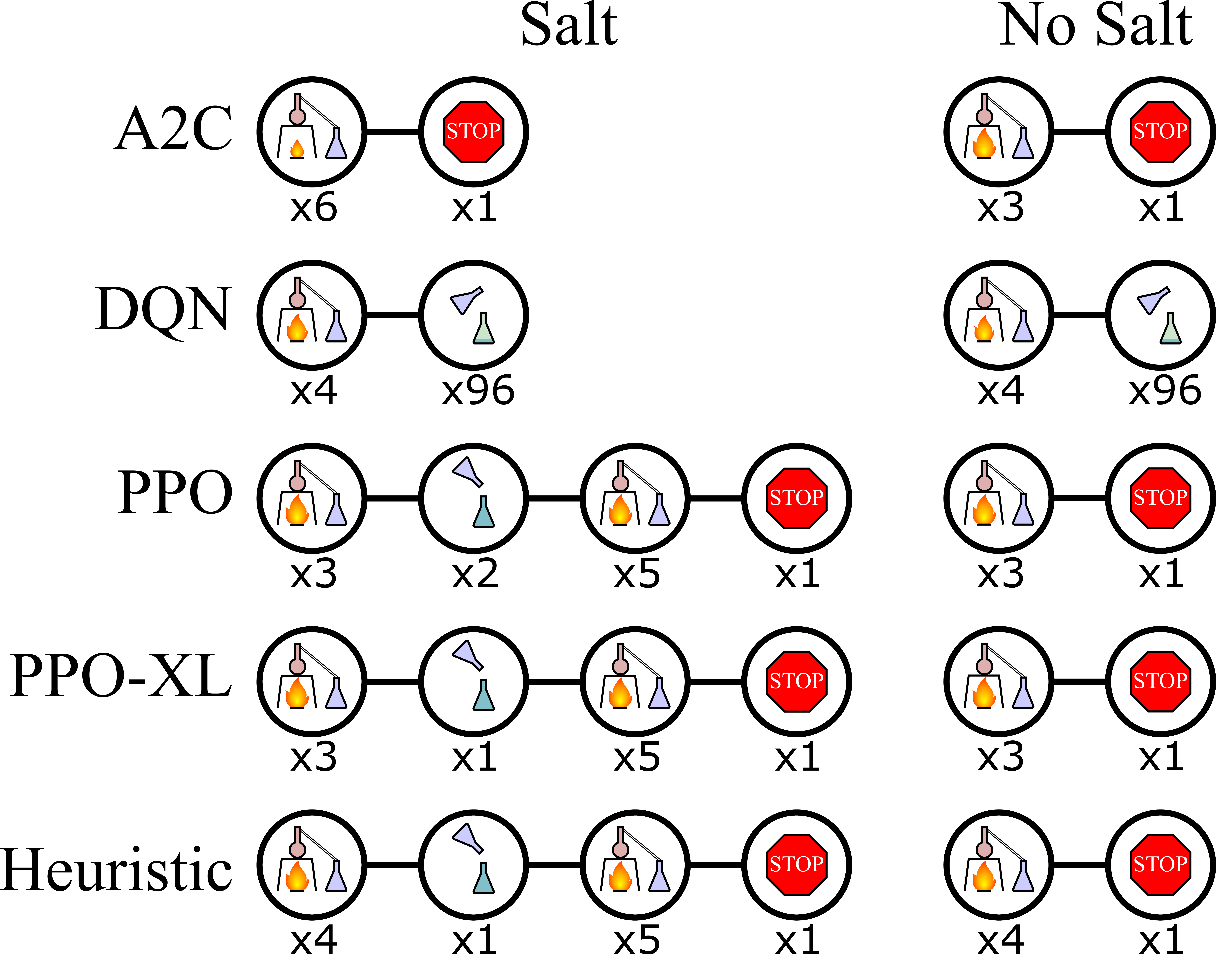}
    \caption{Wurtz \texttt{DiT}, the sequences of actions with the highest return produced by the best performing policy learned with each algorithm for two cases: when salt is and is not initially present in the distillation vessel with another material. Each picture represents an action and average value described by Fig. \ref{fig:distill_actions}. The number beneath the image represents how many times that action was repeated. The PPO, PPO-XL, and heuristic policies are nearly identical in both cases, with only minor differences. When no salt is present, the A2C and DQN policies are similar to the others, however when salt is present they continue to behave as if it is not.}
    \label{fig:wurtzdistill_policy}
\end{figure}

For the case when no salt is present in the distillation vessel, the the best performing agents trained with each algorithm learn a very similar policy to the heuristic one, as seen in Fig. \ref{fig:wurtzdistill_policy}. They heat the vessel until the solvent has boiled away, then end the experiment. The agent trained with DQN however, was the only one not to manually end the experiment but rather repeat an action to the point it is equivalent to waiting until the maximum number of steps has elapsed. For the case when salt and an additional material are present, the agent trained with A2C learned to only slightly modify its actions compared to the previous case, where the one trained with DQN makes no modifications. The best performing agents trained with PPO and PPO-XL modify their actions similar to the heuristic policy, achieving the optimal return in both cases. This shows that the expected behavior in our final bench can also be learned by an RL agent.

\section{Limitations}

Currently ChemGymRL has many limitations; any reaction or material that one wishes to model must be predefined with all properties specified by the user.  Additionally, the solvent dynamics are modeled using simple approximations and while they suffice for these introductory tests, they would not for real-world chemistry. 

As previously mentioned, the ChemGymRL framework was designed in a modular fashion for the ease of improvement. The differential equations used to model the reactions could be replaced with a molecular dynamics simulation. This would allow \texttt{RxN} to operate with on a more generalizable rule-set. Without having to manually define the possible reactions, the \texttt{RxN} could be used to discover new, more efficient reaction pathways by an RL agent. Currently, the reward metric used in \texttt{RxN} is the molar amount of desired material produced by the agent. If this metric was changed to reflect a  certain desired property for the produced material, then the \texttt{RxN} could be used for drug discovery. Making similar improvements to \texttt{ExT} and \texttt{DiT}, the RL agent could then learn to purify these new discoveries.

\section{Future Work}
For future work on ChemGymRL, the lab manager environment will be formatted in a way that allows an RL agent to operate in it. Using pre-trained agents for the individual benches, the lab manager agent would decide which vessel to give to which bench while also specifying the desired goal to each bench, in order to achieve the lab manager's own goal. The lab manager agent would make proper use of the agentless characterization bench introduced here as well. In addition to this, the implementation of new benches will be explored, allowing more complicated experiments to be conducted.

\section{Conclusions}

We have introduced and outlined the ChemGymRL interactive framework for RL in chemistry. We have included three benches that RL agents can operate and learn in. We also include a characterization bench, for making observations, which agents do not operate in, and presented directions for improvement. To show these benches are operational, we have successfully, and reproducibly, trained at least one RL agent on each of them. Included in this framework is a vessel state format compatible with each bench, therefore allowing the outputs of one bench to be the input to another.

In the Wurtz \texttt{RxN} experiment, A2C, SAC, and TD3 were not able to show better performances than an agent taking random actions, where PPO was able to achieve optimal returns on all targets. In the second \texttt{RxN} experiment, A2C, SAC, and TD3 were able to show performances that achieves optimal returns for one of the two difficult tasks, whereas PPO was able to achieve optimal returns on both.

In the Wurtz \texttt{ExT} experiment, A2C and DQN were not able to produce agents that perform better than doing nothing, whereas PPO was able to achieve higher returns than the devised heuristics. In the Wurtz \texttt{DiT} experiment each algorithm was able to produce an agent that performs better than doing nothing and much better than an agent taking random actions.

Finally, we included elaborate discussions on other RL algorithms that can be tried in ChemGymRL and how this environment will be extremely valuable to the broader RL research community. We hope to see a wide adoption of ChemGymRL within the set of testbeds commonly used by researchers in the RL community.

\section*{Acknowledgements}

C. Beeler and C. Bellinger performed work at the National Research Council of Canada under the AI4D program. I.T. acknowledges NSERC.

\bibliography{chemgymrl-intro}
\bibliographystyle{synsml2023}

\newpage
\section{Appendix}

\subsection{RL Testbed}\label{sec:rltestbed}

From the perspective of a RL researcher, ChemGymRL provides a very useful training environment on a problem having real-world impact. While majority of prior RL environments focus on games, researchers now acknowledge that RL has become a mature technology that can be useful in a variety of real-world applications \cite{dulac2021challenges}. RL is growing rapidly and many new sub-fields within RL have emerged over the last five years \cite{sutton2018reinforcement}. In this section we highlight the specific RL areas where we think ChemGymRL will be helpful, with the objective of encouraging the RL community to adopt this library within their suite of RL training environments.

Tab. \ref{tab:algorithms} captures a set of RL paradigms (and associated algorithms) where we believe ChemGymRL will be impactful. In the previous sections, we considered a small set of RL algorithms to benchmark performances in ChemGymRL. However, there are a much larger class of RL sub-fields where ChemGymRL can be used as a testbed. 

\begin{table*}[ht]
\fontsize{8pt}{8pt}\selectfont
\centering
\setlength{\tabcolsep}{6pt} 
\renewcommand{\arraystretch}{1.5} 
\setlength\arrayrulewidth{1.5pt}
    \begin{tabular}{|p {0.20\linewidth}|p {0.20\linewidth}|p {0.20\linewidth}|p{0.20\linewidth}|}
    \hline 
    \textbf{Paradigm} & \textbf{Algorithm} & \textbf{Other Testbeds} & \textbf{ChemGymRL Application} \\ \hline
                                                 & I2A \cite{racaniere2017imagination}  & Sokoban \cite{shoham2021solving} &  \\ \cline{2-3} 
    \multirow{-2}{3.4cm}{Model-based RL \cite{moerland2023model}} & ME-TRPO \cite{kurutach2018model}  & Mujoco \cite{todorov2012mujoco} & \multirow{-2}{3.4cm}{Learn a model for chemical reactions and plan using the model} \\ \hline
                                                 & OOCG \cite{silva2018object} & Half Field Offense \cite{hausknecht2016half} & \\ \cline{2-3} 
                                                 & ScreenNet \cite{kim2018screenernet} & Minecraft \cite{guss2019minerl} &  \\ \cline{2-3} 
                                                 & H-DRLN \cite{tessler2017deep} & StarCraft-II \cite{samvelyan2019starcraft} & \\ \cline{2-3} 
    \multirow{-6}{3.4cm}{Curriculum Learning \cite{narvekar2020curriculum}}        & CM3 \cite{yang2018cm3} & SUMO \cite{lopez2018microscopic} & \multirow{-6}{3.4cm}{Generate a curriculum of tasks across benches} \\ \hline
                                                 & RCPO \cite{tessler2018reward}  & Mujoco \cite{todorov2012mujoco} &  \\ \cline{2-3} 
    \multirow{-3}{3.4cm}{Reward Shaping \cite{laud2004theory}} & DPBA \cite{harutyunyan2015expressing}  & Cartpole \cite{sutton:1998} & \multirow{-3}{3.4cm}{Provide shaping rewards for intermediate steps} \\ \hline
                                                 & DRQN \cite{hausknecht2015deep}  & Atari \cite{Mnih2015} &  \\ \cline{2-3} 
    \multirow{-2}{3.4cm}{Partial Observability \cite{graves2016hybrid}} & DNC \cite{kurutach2018model}  & POPGym \cite{morad2023popgym} & \multirow{-2}{3.4cm}{True state not observed as in extraction bench} \\ \hline
                                                 & C51 \cite{bellemare2017distributional}  & Atari \cite{Mnih2015} &  \\ \cline{2-3} 
    \multirow{-2}{3.4cm}{Distributional RL \cite{bellemare2017distributional}} & QR-DQN \cite{dabney2018distributional}  & Windy Gridworld \cite{sutton:1998} & \multirow{-2}{3.4cm}{Policy based on distribution of reagents produced for every action} \\ \hline
                                                 & HER \cite{andrychowicz2017hindsight}  & Montezuma's Revenge \cite{Mnih2015} &  \\ \cline{2-3} 
    \multirow{-2}{3.4cm}{Experience Replay Methods \cite{fedus2020revisiting}} & ERO \cite{zha2019experience}  & Mujoco \cite{todorov2012mujoco} & \multirow{-2}{3.4cm}{Sparse rewards only obtained at the end of the episode} \\ \hline
                                                 & DQfD \cite{hester2018deep}  & Atari \cite{Mnih2015} &  \\ \cline{2-3} 
    \multirow{-2}{3.4cm}{Learning from Demonstations \cite{piot2014boosted}} & NAC \cite{gao2018reinforcement}  & GTA-V \cite{gao2018reinforcement} & \multirow{-2.5}{3.4cm}{Demonstrations from hueristic/human policies} \\ \hline
                                                 & Option-critic\cite{bacon2017option}  & Atari \cite{Mnih2015} &  \\ \cline{2-3} 
    \multirow{-2}{3.4cm}{Hierarchical RL \cite{sutton1999between}} & AOC \cite{chunduru2022attention}  & Four Rooms domain \cite{sutton1999between} & \multirow{-2}{3.4cm}{Hierarchical policy in terms of benches (options) and low-level actions} \\ \hline
                                                 & IPO \cite{liu2020ipo}  & Safety Gym \cite{ray2019benchmarking} &  \\ \cline{2-3} 
    \multirow{-2}{3.4cm}{Constrained RL \cite{ray2019benchmarking}}               & CPO \cite{achiam2017constrained}  & Mujoco \cite{todorov2012mujoco} & \multirow{-2}{3.4cm}{ Safe handing of chemical reagents} \\ \hline
    \end{tabular}
\caption{A non-exhaustive list of RL paradigms and algorithms that are potentially applicable to ChemGymRL.} 
\label{tab:algorithms}
\end{table*}

Traditionally most RL methods are model-free, where the transition dynamics is neither known nor learned by the algorithms \cite{sutton2018reinforcement}, and most algorithms we considered in this paper fall into the model-free category. Alternatively, another class of algorithms combines planning \cite{russell2010artificial} and RL. These algorithms explicitly learn the transition dynamics model and then use this model to arrive at the optimal policy (by planning). Such algorithms are classified as model-based RL algorithms \cite{moerland2023model}. Depending on the domain, model-based RL methods can be significantly more sample efficient than their widely used model-free counterparts \cite{wang2019benchmarking}. In ChemGymRL, model-based approaches that learn the transition dynamics can effectively plan over all the different possible ways of obtaining higher amounts of the target products from the input reactants.

Another class of methods applicable to ChemGymRL is the curriculum learning methods in RL \cite{narvekar2020curriculum}. In this paradigm, RL problems that are too hard to learn from scratch can be broken down into a \emph{curriculum} of tasks which can then be tackled individually. Curriculum learning methods generally have 3 key elements: task generation, sequencing, and transfer learning. Task generation is the process of generating a good set of tasks that are neither trivial nor too hard to solve. Sequencing is the process of generating a sequence (in terms of difficulty, cost etc.) of available tasks. Transfer learning focuses on strategies to transfer knowledge from one task to another (so that the agent does not have to learn each task from scratch). Generating desired products from available reactants is a complex process that requires learning policies across multiple benches in ChemGymRL. This renders itself well to the curriculum learning framework.  Closely related to curriculum learning is the hierarchical RL approach using the options framework \cite{sutton1999}. Transferring higher-level knowledge across tasks can take the form of partial policies of options. Options can be seen as temporally extended actions which allow learning/planning over a sequence of lower-level actions.

The default reward function in ChemGymRL is sparse. For example, in the \texttt{RxN} the agent only receives a reward equal to the molar amount of the target material produced at the end of the episode. Reward shaping methods \cite{laud2004theory} provide small rewards in intermediate steps to help the agent learn and converge faster. It is possible for reward shaping to change the optimal behaviour and make agents learn unintended policies. Potential-based reward shaping methods are a special class of reward shaping methods that preserves the optimality order over policies and does not affect the converged optimal behaviour of the MDP \cite{ng1999policy, wiewiora2003principled}. Such methods can be used in ChemGymRL. 
 
The benches in ChemGymRL are typically partially observable. For example, the \texttt{ExT} had (partial) observations that do not show the amount of dissolved solutes present nor their distribution throughout the solvents. In this paper, we have considered benchmarks that assume that the observation obtained is the true state (such methods perform surprisingly well in partially observable domains too, in many cases \cite{hausknecht2015deep}). Alternatively, using other algorithms such as DRQN \cite{hausknecht2015deep} or DNC \cite{kurutach2018model} that explicitly reason over partial observations, is guaranteed to provide better empirical performances. 

RL algorithms traditionally aim to maximize the \emph{expected} utility using the $Q$-function or the value function. However, in many environments considering the entire distribution of returns rather than just the expected value has been demonstrated to be helpful \cite{bellemare2017distributional}. This distributional perspective of RL has gained a lot of attention recently in the RL research community \cite{bellemare2023distributional}. Distributional algorithms can show good performances in the ChemGymRL environments since the returns are typically multi-modal (in the sense of distributions) with more than one product expected to be produced as a result of chemical reactions.  Hence, these methods can also use ChemGymRL as a testbed. 

The commonly used RL algorithms such as DQN \cite{Mnih2015} and Deep Deterministic Policy Gradients DDPG \cite{lillicrap2015continuous} use an experience replay technique for improving sample efficiency and removing correlations between successive data samples. Recently, there have been many advances in RL algorithms that aim to improve the efficiency of the experience replay technique, especially in sparse reward settings. Several such algorithms, including the popular human experience replay (HER) \cite{andrychowicz2017hindsight} and experience replay optimization (ERO) \cite{zha2019experience} methods can be tested in ChemGymRL. 

Another paradigm of interest is the \emph{Learning from Demonstrations} (LfD), which combines imitation-learning from external demonstrations with RL approach of learning from the environment \cite{piot2014boosted}. The objective is to make use of pre-existing policies or human knowledge to accelerate the training of RL agents, as opposed to learning from scratch which is highly sample inefficient. In ChemGymRL, for all the benches we have provided hueristic policies that can be used as the external demonstrator to train RL algorithms. Popular LfD techniques like deep $Q$-learning from demonstrations (DQfD) \cite{hester2018deep} and normalized actor-critic (NAC) \cite{gao2018reinforcement} are of interest in ChemGymRL. 

Finally, while handing chemical reagents, it is not only sufficient to learn the most optimal policy that provides higher quantities of the required products but also use safe techniques that do not cause any harm or injuries to agents or humans using these reagents. Such constraints in learning the optimal behaviour can be principally incorporated within the \emph{Constrained RL} framework \cite{ray2019benchmarking}. The constrained RL techniques like interior-point policy optimization (IPO) and constrained policy optimization (CPO) \cite{achiam2017constrained} impose constraints in the MDP that curtail an agent's ability to explore, so that it can perform safe exploration \cite{garcia2015comprehensive}. Such methods are also expected to be successful in ChemGymRL.  

We would like to highlight in Tab. \ref{tab:algorithms} that many of the other testbeds used by prior works pertained to computer games, video games, or robotic simulations. In this context, ChemGymRL provides an excellent alternative testbed that pertains to a real-world problem and helps in evaluating the performance of a large class of RL algorithms. 

\subsection{Reactions}
As an example, consider the reaction $X + Y \rightarrow Z$. Its system of differential equations is defined as
\begin{equation}
    \begin{aligned}
        \frac{\partial [X]}{\partial t} &= -k[X][Y] \\
        \frac{\partial [Y]}{\partial t} &= -k[X][Y] \\
        \frac{\partial [Z]}{\partial t} &= k[X][Y],
    \end{aligned}
\end{equation}
where $[X]$ denotes the concentration of material $X$ and $k$ is the rate constant, defined by the Arrhenius equation
\begin{equation}
    k = Ae^{\frac{E_{a}}{RT}},
\end{equation}
where $A$ is the pre-exponential factor, $E_{a}$ is the activation energy, $R$ is the ideal gas constant, and $T$ is the temperature. The possible reactions that can occur in this bench are determined by selecting a family of reactions from a directory of supported, available reactions. New reactions can be easily added to this bench by following the provided template. This reaction template includes parameters for the pre-exponential factors, and the activation energies, the stoichiometric coefficients of each material for each reaction.

Chemical reactions of this form can be considered as special cases of the initial value problem:

\begin{equation}
    \begin{aligned}
        \frac{d\vec{y}}{dt} = \vec{f}(t,\vec{y}),
    \end{aligned}
\end{equation}
where $\frac{\partial\vec{f}}{\partial t} = \vec{0}$. Note, $\vec{y}$ are your concentrations and $\vec{f}(t,\vec{y})$ are your rates. The RK45 (Runge-Kutta-Fehlberg) method \cite{hairer1993solving} was used to solve these ODE equations and obtain new chemical concentrations as time passes.

\subsection{Extractions}
In the layer separation equations we present here, we consider the solvents settling as moving forward in time and mixing the contents of a vessel as moving backwards through time. This \texttt{ExT} uses Gaussian distributions to represent the solvent layers. For solvents $L_{1}$, $L_{2}$, $\ldots$, $L_{n}$, the center of solvent $L_{i}$, or mean of the Gaussian, is given by
\begin{equation}
    \mu_{L_{i}} = (t - t_{\text{mix}})\sum_{j=1, j \neq i}^{n} (D_{L_{j}} - D_{L_{i}}),
\end{equation}
where $t_{\text{mix}} \leq t$ is the time value assigned to a fully mixed solution and $D_{L_{i}}$ is the density of $L_{i}$. The spread of solvent $L_{i}$, or the variance of the Gaussian, is given by
\begin{equation}
    \sigma^{2}_{L_{i}} = \frac{1}{\sqrt{2\pi}}e^{-t}.
\end{equation}
While these solvents separate, the solutes are being dispersed based on their relative polarities and amounts of the solvents. In the \texttt{ExT}, the relative solute amount at time $t$ in solvent $L$ is defined by
\begin{equation}
    S_{L, t} = \begin{cases} 
      S_{L}^{\ast}(t) + \frac{t - t_{\text{mix}}}{t' - t_{\text{mix}}}(S_{L, t'} - S_{L}^{\ast}(t')) & t < t' \\
      S_{L}^{\ast}(t) + S_{L, t'} - S_{L}^{\ast}(t') & t > t' \\
      S_{L, t'} & t = t'
   \end{cases},
\end{equation}
with $t' = t - \Delta t$ and
\begin{equation}
    \begin{aligned}
        &S_{L}^{\ast}(t) = \left([S]\frac{[L]}{\sum_{l \in \mathcal{L}} [l]}\right)e^{30(t_{\text{mix}} - t)} + \\
        &\quad \frac{1 - \frac{|P_{S} - P_{L}|}{\sum_{l \in \mathcal{L}}|P_{S} - P_{l}|}}{1 - \frac{\sum_{l \in \mathcal{L}} [l]|P_{S} - P_{l}|}{(\sum_{l \in \mathcal{L}} [l])(\sum_{l \in \mathcal{L}}|P_{S} - P_{l}|)}}\left(1 - e^{30(t_{\text{mix}} - t)}\right),
    \end{aligned}
\end{equation}
where $[X]$ is the total concentration of $X$ in the vessel, $\mathcal{L}$ is the set of all solvents in the vessel, and $P_{X}$ is the polarity of $X$.

\end{document}